\documentclass[runningheads]{llncs}
\usepackage{graphicx}
\usepackage{amsmath,amssymb} 
\usepackage{color}
\usepackage{multirow} 
\usepackage{graphicx}
\usepackage{subcaption}
\usepackage{siunitx}
\usepackage{wrapfig}
\usepackage{url}
\usepackage{hyperref}
\begin{document}
\pagestyle{headings}
\mainmatter

\def\ACCV20SubNumber{07}  

\title{BdSL36: A Dataset for Bangladeshi Sign Letters Recognition} 
\titlerunning{BdSL36}
\authorrunning{Oishee Bintey Hoque et al.}

\author{Oishee Bintey Hoque\inst{1}\and
Mohammad Imrul Jubair\inst{2} \and
Al-Farabi Akash\inst{3} \and
Md. Saiful Islam\inst{4}}

\institute{Ahsanullah University of Science and Technology, Bangladesh \\ \email{\{\inst{1}bintu3003, \inst{3}alfa.farabi,\inst{4}saiful.somum\}@gmail.com}\\
\email{\inst{2}mohammadimrul.jubair@ucalgary.ca}}

\maketitle

\begin{abstract}
Bangladeshi Sign Language (BdSL) is a commonly used med\-ium of communication for the hearing-impaired people in Bangladesh. A real-time BdSL interpreter with no controlled lab environment has a broad social impact and an interesting avenue of research as well. Also, it is a challenging task due to the variation in different subjects (age, gender, color, etc.), complex features, and similarities of signs and clustered backgrounds. However, the existing dataset for BdSL classification task is mainly built in a lab friendly setup which limits the application of powerful deep learning technology. In this paper, we introduce a dataset named BdSL36 which incorporates background augmentation to make the dataset versatile and contains over four million images belonging to 36 categories. Besides, we annotate about $40,000$ images with bounding boxes to utilize the potentiality of object detection algorithms. Furthermore, several intensive experiments are performed to establish the baseline performance of our BdSL36. Moreover, we employ beta testing of our classifiers at the user level to justify the possibilities of real-world application with this dataset. We believe our BdSL36 will expedite future research on practical sign letter classification. We make the datasets and all the pre-trained models available for further researcher.
\end{abstract}

\section{Introduction}

Sign language is a non-verbal form of communication used by deaf and hard of hearing people who communicate through bodily movements especially with fingers, hands and arms. Detecting the signs automatically from images or videos is an appealing task in the field of Computer Vision. Understanding what signers are trying to describe always requires recognizing of the different poses of their hands. These poses and gestures differ from region to region, language to language, i.e. for American Sign Language (ASL) \cite{Ye_2018_CVPR_Workshops}, Chinese Sign Language (CSL) \cite{8719155}, etc. A considerable number of people in Bangladesh rely on BdSL \cite{8554466} to communicate in their day to day life \cite{tarafder_akhtar_zaman_rasel_bhuiyan_datta_2015}, and the need for a communication system as a digital interpreter between signers and non-signers is quite apparent. Due to the current resonance of Deep Learning, hand pose recognition \cite{hp1,hp2} from images and videos have advanced significantly. In contrast, sign letter recognition through a real-time application has received less attention---especially when it comes to Bangladeshi Sign Language. In this work, we tend to simplify the way non-signers communicate with signers through a computer vision system for BdSL by exploiting the power of data and deep learning tools. 

Most of the previous work on BdSL and other sign languages \cite{8237594,con1,con2,con3} can be described as a traditional machine learning classification framework or as a mere research as an academic purpose where less focus has been given to build a real-time recognizing system. We can generalize these researches in two modules: (1) datasets with several constraints (i.e. showing only hands, single-colored background) or (2) traditional machine learning classifiers with handcrafted features. These constrained feature-based methods are not ideal for real-life applications or to build a proper sign interpreter as these methods rely on the careful choice of features. With a real-world scenario, the classifier may fail to distinguish the signs from a clustered background.
\begin{figure}
\includegraphics[width = 120mm]{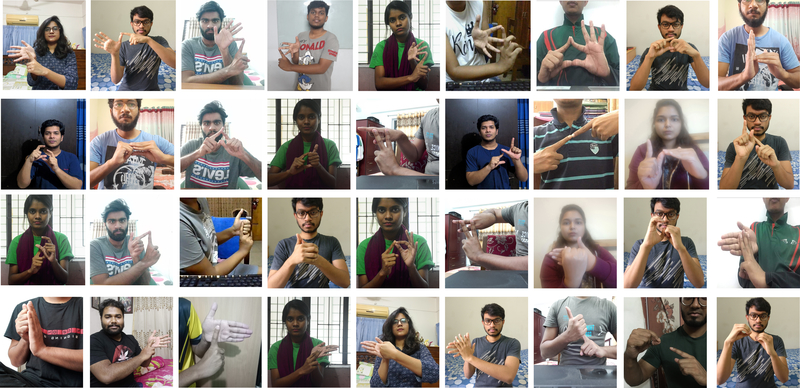}
\caption{Example images of initially collected BdSL36 dataset. Each image represents different BdSL sign letter. Images are serially organized according to their class label from left to right.}
\label{fig1}
\end{figure}

In this recent era, deep learning enables the scope of overcoming these constraints and achieve a state-of-the-art performance on a variety of image classification tasks.  But, to train a deep learning classifier, the role of a large dataset is unavoidable and plays a pivotal role to build a robust classifier. However, deep learning methods, so far, on BdSL letters recognition are restricted to constraint or small scale dataset \cite{fuzzy,contouranalysis,automatic}, which is not useful to build a robust classifier. The collection of sign letter data is challenging due to the lack of open-source information available on the internet, human resources, the deaf community and knowledge. On the other hand, BdSL letters may have high appearance similarity, which might be confusing to human eyes as well. Hence, with handcrafted features or small scale datasets, it is a tough job to accomplish to build a robust interpreter.

To advance the BdSL letters recognition research in computer vision and to introduce a possible real-time sign interpreter to the community in need, we present the BdSL36, a new large-scale BdSL dataset in this work. As mentioned earlier, many factors hinder the process of collecting images with variation to make a large scale sign letter dataset. We organize $10$ volunteers of different ages and gender to help us with the whole process of collecting the images. Initially, we receive $1200$ images that are captured by the volunteers at their convenient and natural environment (see fig. \ref{fig1}) followed by thorough checking by the BdSL experts to assure relevancy of the signs in the data. However, only $25$ to $35$ images per class for building a multi-class classifier of $36$ classes are not enough. To compensate for the low data quantity issue, data augmentation has become an effective technique to improve the accuracy of modern image classifiers by increasing both the amount and diversity of data by randomly augmenting \cite{Cubuk_2019_CVPR,aug2}. The most common augmentation techniques are translation by pixels, rotation, flipping, etc. Intuitively, data augmentation teaches a model about invariances, making the classifier insensitive to flips and translations. Subsequently, to incorporate potential invariances, data augmentation can be easier than hard-coding them into the model directly. In case of a dataset of hand poses, we need variation in hand shape, skin color, different postures of the sign, etc. Moreover, making the dataset insensitive to the background, it is also essential to introduce background variation with the subject being in a different position in the images. To the best of our knowledge, less attention has been paid to find better data augmentation techniques that incorporate more invariances and resolve the background variation limitation in a small scale dataset.

In our work, initially, we incorporate traditional data augmentation---rotation, translation, brightness, contrast adjustment, random cropping, zooming, shearing, cutout \cite{cutout}, etc---on our raw collected dataset which we called BdSL36 version-1 (BdSL35v1). Though these augmentations improve accuracy on the validation set, when we deploy the classifier and test it with real-life users, it performs poorly. Therefore, we employ a new augmentation technique---not widely used as per our knowledge---the background augmentation to introduce second version (BdSL36v2). Here, we removed the background from our raw dataset images, perform traditional augmentation on them, and later, stack them over one million background images downloaded from the internet at random positions. After that, we perform another set of suitable traditional augmentation (perspective and symmetric warp with random probability) and build a dataset of over four million images. Our intensive experiments show that this method achieves excellent improvement in terms of both accuracy and confidence rate. We again perform another set of beta testing with two of the classifiers---trained with background augmented and non augmented data---and achieve a significant improvement in the results. This experiment shows, even with a very small scale dataset, it is possible to get the state-of-the-art performance for complicated features and surpass the limitation of the quantity of the dataset. With this dataset, we also present another version of the dataset (BdSL36v3) with bounding box labeling for object detection methods which contains over $40$ thousand images labeled manually. Our intention in these experiments is to help the deaf community in real life by not limiting the sign letter recognition task to a controlled environment. To further validate the value of our proposed dataset, we also report evaluation on different deep learning networks for the state-of-the-art classification.\\
It is worth to note that, in some of the similar papers, the term 'fingerspelling' is used to refer to the letters of a sign language system, while many other works represent it as 'sign language'. To minimize the ambiguity, in our paper, we use the term 'sign letters' in such context; and the term 'BdSL letters' to refer the letters of Bangladeshi sign language in particular.

Our contributions are summarized as follows:
\begin{itemize}
\item To best of our knowledge, we build the first largest dataset of the Bangladeshi Sign Language letter---\textbf{BdSL36}---for deep learning classification and detection. Our dataset comes with four different versions: raw dataset (\textbf{BdSL36x}), augmentation on the raw dataset (\textbf{BdSL36v1}), background augmented images (\textbf{BdSL36v2}) and bounding box labelled dataset (\textbf{BdSL36v3}). We make all of these versions available to the community so that researchers can exploit them according to their requirements. We incorporate background augmentation technique for small scale dataset with the proof of significant improvement in results through extensive experiment. Dataset can be found  at \href{https://rb.gy/mm3yjg}{\texttt{rb.gy/mm3yjg}}.
\item We conduct extensive experiments on the datasets using deep learning classification and detection models to establish the baseline for future research. We also perform beta testing to identify the possibility of deploying this system in the real world. All these work and experiments are made available to the research community for further investigations.
\end{itemize}

\section{Related Works}
\label{sec:relatedWork}
In this section, we first discuss some of the previous works and datasets in automating BdSL letters recognition, followed by reviewing some works on data augmentation.

Rahaman et al. \cite{contouranalysis} use $1800$ contour templates for 18 Bangladeshi signs separately. The authors introduce a computer vision-based system that applies contour analysis and Haar-like feature-based cascaded classifier and achieve $90.11\%$ recognition accuracy. Ahmed and Akhand \cite{fingertipposition} use 2D images to train an artificial neural network using tip-position vectors to determine the relative fingertip positions. Though they claim to have an accuracy of $98.99\%$ in detecting the BdSL, their approach is not applicable for real-time recognition. M. A. Rahaman et al. in \cite{fuzzy}, present a real-time Bengali and Chinese numeral signs recognition system using contour matching. The system is trained and tested using a total $2000$ contour templates separately for both Bengali and Chinese numeral signs from 10 signers and achieved recognition accuracy of $95.80$\%  and $95.90$\%  with a computational cost of $8.023$ milliseconds per frame. But these classifiers only work in a controlled lab environment. In \cite{automatic}, a method of recognizing Hand-Sign-Spelled Bangla language is introduced. The system is divided into two phases -- hand sign classification and automatic recognition of hand-sign-spelled for BdSL using the Bangla Language Modeling Algorithm (BLMA). The system is tested for BLMA using words, composite numerals and sentences in BdSL achieving mean accuracy of $93.50$\%, $95.50$\% and $90.50$\% respectively. In \cite{FasterRCNN}, the authors use the Faster R-CNN model to develop a system that can recognize Bengali sign letters in real-time and they also propose a dataset of $10$ classes. They train the system on about $1700$ images and were able to successfully recognize $10$ signs with an accuracy of $98.2\%$.

The available BdSL datasets are not sufficient enough \cite{fuzzy,contouranalysis,fingertipposition,FasterRCNN,9019931}  to develop a fully functioned real-time BdSL detection system. Half of these datasets \cite{fuzzy,contouranalysis,fingertipposition,9019934} are also not available for further research. These datasets are further built on a controlled lab environment and the only exception is \cite{FasterRCNN} but this dataset only contains $10$ classes.

Recently, data augmentation in deep learning technology widely attracts the attention of researchers. In \cite{Cubuk_2019_CVPR} introduces an automated approach to find the right set of data augmentation for any dataset through transferring learned policies from other datasets. Other auto augmentation techniques by merging two or more samples from the same class are proposed in \cite{SmartAugmentation,DatasetAugmentation}. Specific augmentation techniques such as flipping, rotating and adding different kinds of noise to the data samples, can increase the dataset and give better performance \cite{Effectiveness}. For generating additional data, adversarial networks have also been incorporated \cite{GAN,gan2} to generate direct augmented data.

\section{Our BdSL Dataset}
\label{sec:dataset}

\subsection{Data Collection and Annotation}
We collect and annotate the BdSL36 datasets with the following five stages: 1) Image collection, 2) Raw data augmentation, 3) Background removal and augmentation, 4) Background augmentation, and 5) Data Labeling with Bounding Box.
\subsubsection{Image Collection.}
\label{subsec:imgCollection}
We establish a dataset for real-time Bangla sign letter classification and detection. Firstly, we visit a deaf school and learn about the letters they practically use in their daily life. There are total $36$ Bangla sign letters in total. To build a dataset with no background constraint or controlled lab environment, we make sure to have fair variation in the BdSL36 in terms of subject and background for both classification and detection datasets. At first step we organize 10 volunteers to collect the raw images, and two experts on BdSL train them to perform this task. All of them use their phone cameras or webcam to capture the images at their convenient environment. After the collection of images from the volunteers, each image is checked by BdSL signer experts individually and they filter out the images which contain signs in the wrong style. After filtration, each class contains $25$ to $35$ images with a total of $1200$ images altogether. We have also incorporated BdSlImset \cite{FasterRCNN}, containing $1700$ images for $10$ classes in our  BdSL36 image dataset, which sums up to $2712$ images in total for 36 classes (see fig. \ref{fig1}).

\subsubsection{Raw Image-Data augmentation.}
\begin{figure}
\centering
  \begin{subfigure}[b]{0.45\linewidth}
    \centering
    \includegraphics[width=1\linewidth]{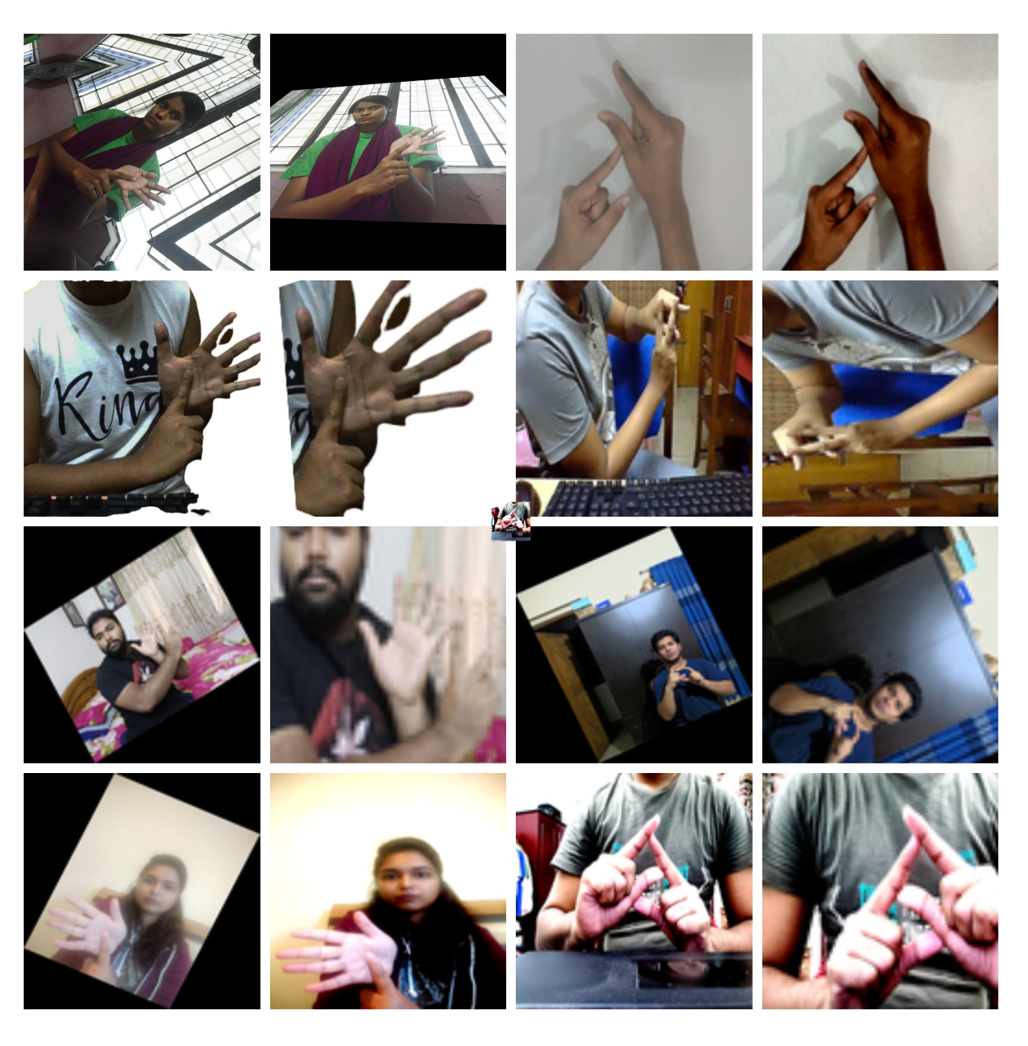} 
    \caption{} 
    \label{dataset1} 
    \vspace{2ex}
  \end{subfigure}
  \begin{subfigure}[b]{0.45\linewidth}
    \centering
    \includegraphics[width=1\linewidth]{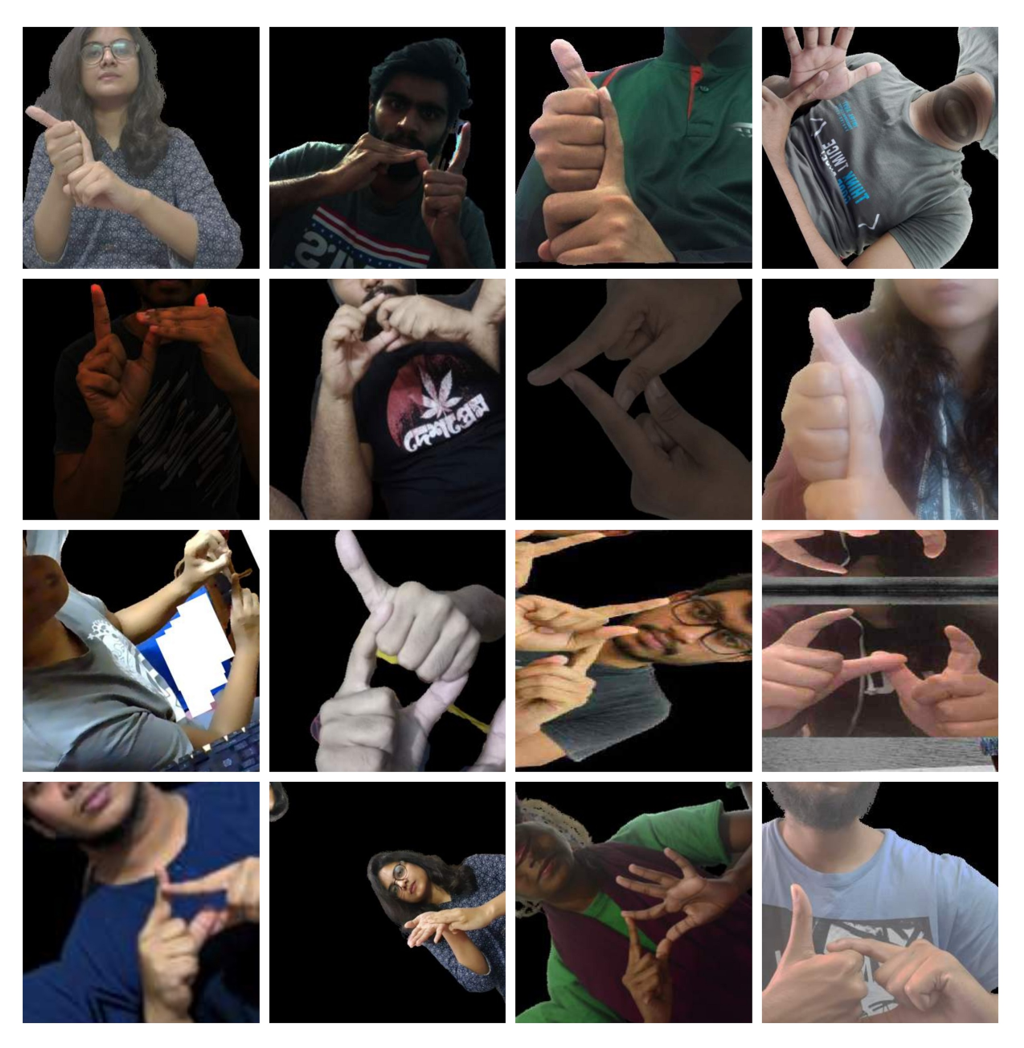} 
    \caption{} 
    \label{dataset2} 
    \vspace{2ex}
  \end{subfigure}

  \caption{(a) Examples of background augmented images from BdSL36x dataset, (b) Examples of bounding box annotated images from BdSL36v3 dataset.}
  \label{fig7} 
\end{figure}

For a real-time sign letter detection, the classifier must recognize the signs accurately in any form with any variation in scale, appearance, pose, viewpoint, and background. For most of the object classification dataset, it is possible to use the internet as the source of collecting the images. However, in the case of BdSL, there are not enough resources, images, or videos, available on the internet for developing a large scale dataset. Besides, no significant deaf community, awareness and privacy issues are some of the factors that hinder the collection of large scale sign datasets. We surmount this problem with data augmentation. Deep learning frameworks usually have built-in data augmentation utilities, but those can be inefficient or lack some required functionality. We have manually augmented our data with all possible variations. These include several types of transformations on each image with random probabilities, i.e., affine transformations, perspective transformations, contrast changes, noise adding, dropout of regions, cropping/padding, blurring, rotation, zoom in/out, symmetric warp, perspective warp, etc.  After the augmentation process, we delete the images if the features get distorted after the augmentation. Finally, \textbf{BdSl36v1} (see fig. \ref{dataset1}) has about  $26,713$ images in total, each class having $700$ images on average.

\subsubsection{Background Augmentation.}
As mentioned earlier, having a large dataset is crucial for the performance of the deep learning models. Recent advances in deep learning models have primarily attributed the quantity and diversity of data gathered in recent years. However, most approaches used in training these models only use basic types of augmentation; less focus has been put into discovering durable types of data augmentation and data augmentation policies that capture data invariances. In this work, we experiment with background augmentation to generate a new robust dataset. Initially, we manually remove the background from each image of the BdSL36x dataset and perform transformations, i.e. rotation ($\pm \ang{60}$), brightness and sharpness adjustment, scaling ($\pm 10\%$), random crop, and zoom in/out, reflection padding, etc. We generate about 12 images from one image (see fig. \ref{dataset2}). Each image is manually checked by our team to discard the distorted images after these transformations. We utilize the internet to collect more than one million background images of various sorts. Then each of the background-removed augmented, and non augmented images are put into five different backgrounds at random positions. After that, to ensure diversity, we employ another set of transformations---such as perspective warp, symmetric warp, random-crop with various sets of magnitudes---on these images. Finally, we have $473,662$ images in the second version of our dataset: \textbf{BdSL36v2}.

\subsubsection{Data Labeling with Bounding Box.}
To utilize the features of the deep learning detection algorithm, we have also generated a dataset with bounding box labeling. Considering the difficulty and cost of labeling, we randomly select some images from BdSLv2 for each class and we appoint two volunteers to label the bounding boxes. Each class contains $1250$ images on average with a total of $45,000$ images for $36$ classes. The images are labeled following the Pascal VOC \cite{pascal-voc-2012} and also a version of YOLO \cite{redmon2018yolov3} format is available.

\begin{figure}
\includegraphics[width = 120mm]{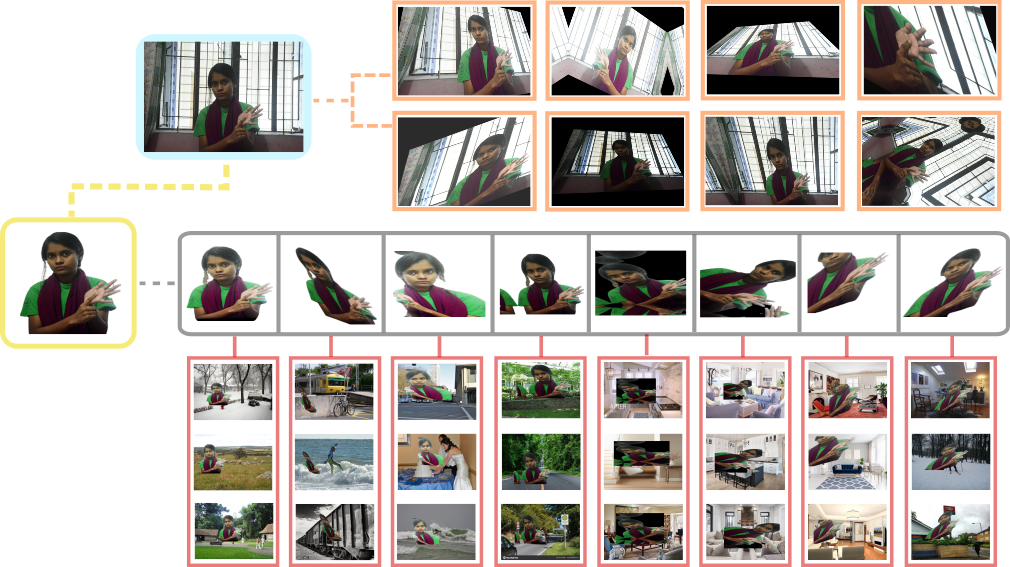}
\caption{The top left image is a raw image captured by one of the volunteers. The two rows to it's right shows the sample augmented images generated from the main image. The leftmost image under the main image has the background removed. Later, augmentation has been applied to this image shown to it's right. From those augmented images, we can see each image contain another set of 3 images with a different background at a different position. All of these images are different from the main image in terms of shape, brightness, contrast, viewpoint background, etc. And, the background augmentation helps to add more variation in a small scale dataset.}
\label{flow}
\end{figure}

A visual representation of the flow of our dataset has been shown in fig. \ref{flow}.

\subsection{Dataset Split}
The BdSL36 dataset contains more than four million images in total. We have three versions of our dataset. The BdSL36v1 contains around 22,000 images and only split into train and validation set following 80:20 split where the train set contains approximately 17,600 and validation set 4,400 images. Subsequently, BdSL36v2 contains around 400,000 images and we follow a roughly 70:5:15 split as training, validation and testing set. Specifically, this dataset is nearly split into 300,000 training, 20,000 validation and 60,000 test images. On the other hand, for the object detection dataset, BdSL36v3 is split into 9:1 as 45,500 training and 4,500 validation images. 

\subsection{Comparison with Other Datasets}
In Table \ref{comparedata}, we compare the BdSL36 with other existing datasets related to the task of BdSL letter recognition. Most of the available BdSL \cite{contouranalysis,FasterRCNN,9019931,9019934,Rahaman2014RealtimeCV,1604d862454045289528f61b056e4864,8554466} datasets have background constraint and only hand pose is shown which are not suitable for the real-time environment applications. Only BdSLImSet \cite{FasterRCNN} has no such constraints but only has 10 classes. Even if we only consider BdSL36x dataset from BdSL36, the number of participants and background variation makes BdSL36x much more diverse and suitable than other available datasets for deep learning models. Besides, only half of these datasets are available. Due to all these limitations, with the most existing datasets, a real-world application is hard to achieve. Finally, BdSL36v3, is the first and only complete object detection dataset for BdSL letter detection.

\setlength{\tabcolsep}{4pt}
\begin{table}[]
\caption{Comparison with existing BdSL datasets.  The Avail, Bg. Const, BBox, Avg., `Y' and `N'  denotes the availability of datasets to the public, Background Constraint, Bounding Box, Average, Yes and No respectively.}
\centering
\begin{tabular}{cccccccc}
\noalign{\smallskip}
\hline
\noalign{\smallskip}
\textbf{Dataset} & \textbf{Year} & \textbf{Class} & \textbf{Avail} & \textbf{\begin{tabular}[c]{@{}c@{}}Bg. \\ Const\end{tabular}} & \textbf{\begin{tabular}[c]{@{}c@{}}BBox\\ Label\end{tabular}} & \textbf{Sample} & \textbf{Avg.} \\   \noalign{\smallskip}
\hline
\noalign{\smallskip}
Rahman et al. \cite{Rahaman2014RealtimeCV}   & 2014          & 10             & N              & Y      & N     & 360             & 36           \\
Rahman et al. \cite{contouranalysis}  & 2015          & 10             & N              & Y      & N     & 100             & 10           \\
Ahmed et al. \cite{1604d862454045289528f61b056e4864}     & 2016          & 14             & N              & Y      & N     & 518             & 37           \\
BdSLImset \cite{FasterRCNN}       & 2018          & 10             & Y              & N      & Y     & 100             & 10           \\
Ishara-Lipi  \cite{8554466}      & 2018          & 36             & Y              & Y      & N     & 1,800            & 50           \\
Sadik et al. \cite{9019931}        & 2019          & 10             & Y              & Y      & N     & 400             & 40           \\
Urme et al.  \cite{9019934}        & 2019          & 37             & N              & Y      & N     & 74,000           & 2,000         \\ \hline
OurDataset       & 2020          & 36             & Y              & N      & Y       & 473,662        & 13,157      \\ \hline

\end{tabular}
\label{comparedata}
\end{table}
\setlength{\tabcolsep}{1.4pt}

\section{Experimental Evaluation}
In this section, we empirically investigate the performance of several deep learning object detection frameworks on the BdSL36 dataset to comprehensively evaluate the performance. Our results show that applying only traditional augmentation on a small scale dataset like BdSl36v1 cannot totally overcome the overfitting problem and perform poorly in a real-time environment as well. Whereas, background augmentation yields significant improvement in real-time environment performance without adding any extra raw data, making the classifier learn the invariances more accurately and insensitive to the background. 

\subsection{Experiment Settings.}
\label{eS}
We use deep learning model architectures - ResNet34 \cite{he2015deep} , ResNet50 \cite{he2015deep} , VGGNet\_19 \cite{simonyan2014deep}, Densenet169 \cite{huang2016densely}, Densenet201 \cite{huang2016densely}, Alexnet \cite{3065386} and Squeezenet \cite{i2016squeezenet} - as our base model. We remove the last layer of the models and concatenate an AdaptiveAvgPool2d, an AdaptiveMaxPool2d, a Flatten layer followed by two blocks of [BN-Dropout-Linear-ReLU] layer. The blocks are defined by the linear filters and dropout probability (0.5) arguments. Specifically, the first block will have several  inputs inferred from the backbone base architecture, and the last one will have outputs equal to the number of classes of the data. The intermediate blocks have many inputs/outputs determined by linear filters (1024-$>$512-$>$36), each block having inputs equal to the number of outputs of the previous block. We also add batch normalization at the end, which significantly improves the performance by scaling the output layer. We train the model in two phases: first, we freeze the base model and only train the newly added layers for two epochs to convert the base model's previously analyzed features into the prediction of our data. Next, we unfreeze the backbone layers to fine-tune the whole model with different learning rates.  For primary training, we use a minibatch size of 64; the learning rate is initialized as 0.003, with a div factor of 25, a weight decay of 0.1, and a momentum of 0.9. Also, to avoid overfitting, we employ a dropout of 0.5. For the second part of the training, we only change the learning rates keeping the other parameters identical.  As all these networks are pre-trained on the large scale Imagenet dataset, the earlier layers already know the basic shapes \cite{layers} and do not require much training. The deeper the network goes, the layers get more class-specific. While training the unfreezing layer, we split our model into a few sections based on the learning rate. We keep the learning rate lower for the initial layers than the deeper layers, initialized between $3e^-3$ and $3e^-4$. We use $224 \times 224$ sized images with default training time augmentation incorporated by the library. The in-depth feature-based experiments are implemented using Pytorch, a fastai library, and performed on Kaggle GPU.
\subsection{Evaluation Metrics.}
The BdSL36 dataset has extremely common similarities among some classes. We apply several metrics evaluation on the validation set for the classification tasks, which include precision, recall, FBeta, accuracy and loss.  The precision (denote as pre) quantifies the correctness of a classifier by the ability not to label a negative class as positive. The recall (denote as rec) measures the number of correctly predicted classes out of the number of the actual classes.
The FBeta (denote as FB) score is the weighted harmonic mean of precision and recall. We utilize Average Precision (AP) for IoU= [.50:.05:.95],[.50],[.75]. We denote AP for different IoU configuration as AP, $ AP^{\frac{1}{2}}$ $ AP^{\frac{3}{4}}$ respectively. With higher IoU, it gets difficult to detect for the system.
\subsection{Classification and Detection with Deep Learning Networks.}
In this section, we evaluate the performance of several deep learning model architectures --- ResNet34 \cite{he2015deep} , ResNet50 \cite{he2015deep} , VGGNet\_19 \cite{simonyan2014deep}, Densenet169 \cite{huang2016densely}, Densenet201 \cite{huang2016densely}, Alexnet \cite{3065386} and Squeezenet \cite{i2016squeezenet} --- on BdSL36v2 dataset.

All of these networks are pre-trained on the ImageNet\cite{imagenet_cvpr09} and fined tuned on the BdSL36v2 dataset with hyper-parameters and settings mentioned in section \ref{eS}. Table \ref{models} shows the classification performance on BdSL36v2 validation set of the deep models. The VGGNet\_19 performs best compared to other models with 99.1\% accuracy. All the models in our experiment are trained with the same parameters and the same number of epochs. It might be possible to use different discrete parameters for other models or train for a longer time to perform better. We can see that ResNet34, ResNet50, VGGNet\_19, Densenet169, Densenet201  perform similarly in terms of all the evaluation matrices whereas Alexnet and Squeezenet perform much poorer than the others. Squeezenet has 41.5\% accuracy with  44.4\% precision and 42.4\% recall which are very low compared to other models. 

We also evaluate BdSL36v3 on several state-of-the-art object detection methods. Faster R-CNN (backbone VGG-16) \cite{ren2015faster} a two-stage based method which detects objects through first sliding the window on a feature map to identify the object. Next, it classifies and regresses the corresponding box co-ordinates. Whereas, one stage based methods, SSD300 (backbone Resnet-50) \cite{Liu_2016} and YOLOv3 (backbone DArknet-53) \cite{redmon2018yolov3} skip the first step of FRCNN and directly regress the category and bounding box position. Table \ref{detectionT} shows that two-stage based FRCNN performs better over the other two networks. For the training of this network, we use the base architecture and parameters of the individual networks.
\setlength{\tabcolsep}{4pt}
\begin{table}[]
\caption{Classification performance of the deep learning classifiers under different evaluation metrics on the BdSL36v2 dataset.}
\centering
\begin{tabular}{llllll}
\noalign{\smallskip}
\hline
\noalign{\smallskip}
Methods & Pre & Rec & FB &  Acc & Loss\\   
\noalign{\smallskip}
\hline  
\noalign{\smallskip}
ResNet34 \cite{he2015deep}    & 98.29 & 98.28 & 98.28 &  98.17 & 0.0592	 \\
\noalign{\smallskip}
ResNet50 \cite{he2015deep}     & 98.83 & 98.79 & 98.8 & 98.71 & 0.0421
 \\\noalign{\smallskip}
VGG19\_bn \cite{simonyan2014deep}   & 99.17 & 99.17 & 99.17  & 99.10 & 0.0284\\\noalign{\smallskip}
Densenet169 \cite{huang2016densely}   & 98.67 & 98.64 & 98.64 &  98.55 & 0.0481\\\noalign{\smallskip}
Densenet201 \cite{huang2016densely}   & 98.70 & 98.65 & 98.66 &  98.56 & 0.0145\\\noalign{\smallskip}
Alexnet \cite{3065386}     & 84.1 & 83.9 & 83.9  & 83.1 & 0.58\\\noalign{\smallskip}
Squeezenet \cite{i2016squeezenet}   & 44.4 & 42.4 & 42.2 &  41.5 & 2.15\\\hline
\end{tabular}
\label{models}
\end{table}
\setlength{\tabcolsep}{1.4pt}
\setlength{\tabcolsep}{4pt}
\begin{table}[]
\caption{Average precision performance of object detection methods under different IoU thresholds.}
\centering
\begin{tabular}{lllll}
\noalign{\smallskip}
\hline
\noalign{\smallskip}
Method & Backbone & AP & $ AP^{\frac{1}{2}}$  & $ AP^{\frac{1}{2}}$  \\  \hline
FRCNN \cite{ren2015faster}      & VGG16         & 46.8  & 81.4  & 36.59  \\
YOLOv3 \cite{redmon2018yolov3}          & ResNet50      & 28.1  & 55.3  & 16.5  \\
SSD300 \cite{Liu_2016}  & VGG16         & 41.2   & 79.61   & 36.53   \\\hline
\end{tabular}
\label{detectionT}
\end{table}
\setlength{\tabcolsep}{1.4pt}

\subsection{Further Analysis}
As our goal is to generate a dataset that can produce an applicable real-time system with the help of a deep learning classifier, we further analyze our classifiers with beta testing. We use two of our classifiers: the first one trained on BdSL36v1 dataset (without background augmentation), the second one trained on BdSL36v2 dataset (with background augmentation), both trained on Resnet50. An evaluation on BdSL36v2 test set, between these two classifiers has been shown in Table \ref{user}. It is to be noted that, the validation accuracy shown in Table \ref{user}, is on individual validation datasets.
\setlength{\tabcolsep}{4pt}
\begin{table}[]
\caption{A comparison between BdSL36v1 \&  BdSL36v2 trained classifier with Test Set.}
\centering
\begin{tabular}{lllll}
\noalign{\smallskip}
\hline
\noalign{\smallskip}
Dataset    & Val Acc. & Acc & Pre & Recall \\
\noalign{\smallskip}
\hline
\noalign{\smallskip}
BdSL36v1 & 98.6     &  36.7  & 37.2  & 36.9    \\
\noalign{\smallskip}
BdSL36v2 & 98.83    & 96.1  & 96.8  & 96.2  \\\hline
\end{tabular}
\label{user}
\end{table}
\setlength{\tabcolsep}{1.4pt}

\begin{figure}
\centering
\includegraphics[width=1\linewidth,height=2.4cm]{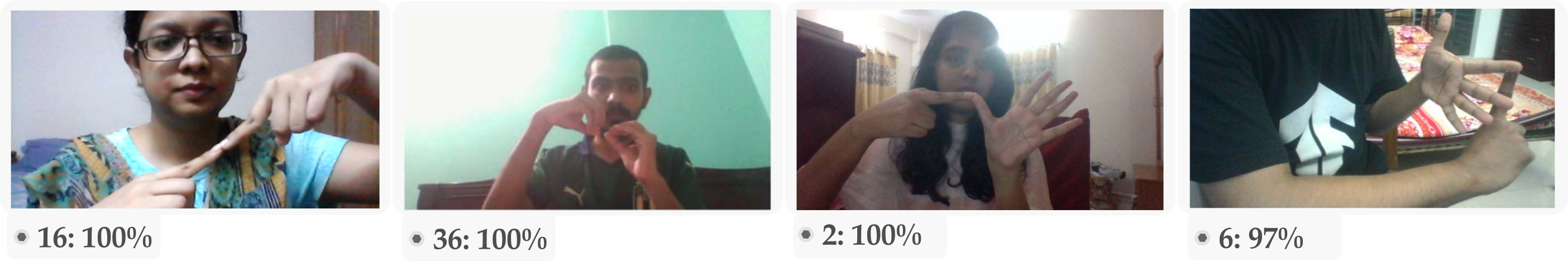}
\caption{Sample images from beta testing with signers with BdSLv2 classifier.}
\label{userfig}
\end{figure}

\begin{figure}%
    \centering
    \subfloat[]{{\includegraphics[width=11cm]{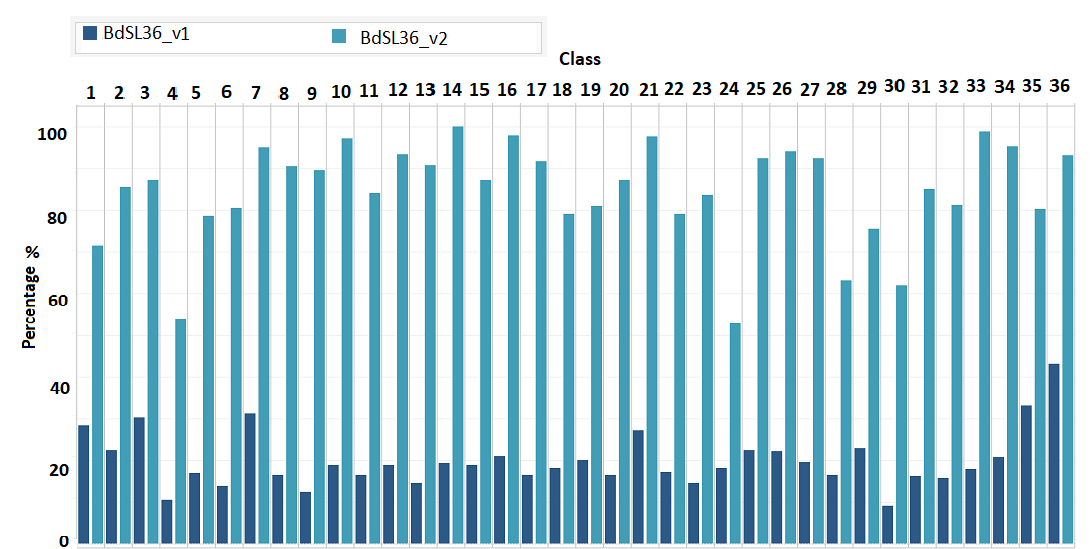} }}%
    \qquad
    \subfloat[]{{\includegraphics[width=11cm]{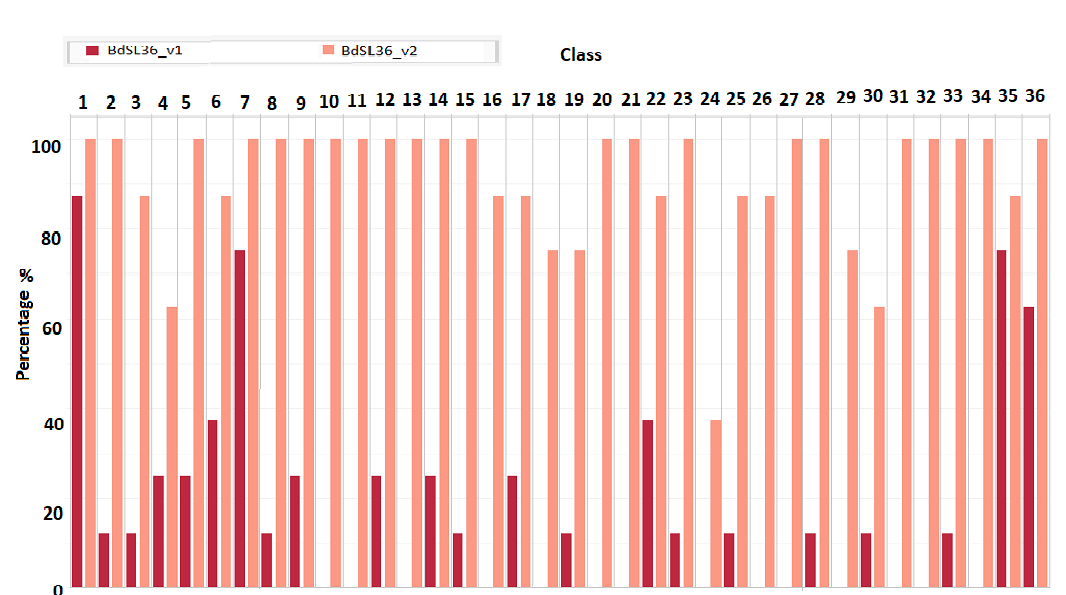} }}%
    \caption{Comparison of (a)confidence rate and (b)accuracy from beta testing between BdSl36v1(without background augmentation) and BdSL36v2(with background augmentation) classifiers.}%
    \label{graph}%
\end{figure}

Though classifier 1 has good accuracy on its validation set, it performs poorly on the test set whereas classier 2 performs significantly well both on the validation set and test set. As the test set is generated from BdSL36v2 dataset, there is a possibility of the result being biased on the tests. So, we run another experiment at the user level to test the robustness of these classifiers in the real-life environment. Eight signers perform this evaluation at their own home, and none of them have participated in the dataset collection process. So, the testing environments and the subjects are new to the system. We ask the users to capture the image of each sign using our system (see fig. \ref{userfig}) and report the prediction values and confidence rate in the provided excel sheet. 
For the correct prediction user inputs 1 or 0 otherwise and the confidence rate of the actual class predicted by the system. For a wrong prediction, we ask the user to capture a sign not more than three times and mention it in the excel sheet.
From the bar chart shown in fig. \ref{graph}, even at the user level, we can see an increase of 60\%  in both accuracy and confidence rate. As the fig. \ref{graph} shows, classifier 1 hardly recognizes any of the signs, and for most of the sign, the users report that they need to capture images multiple times. The confidence rate is admittedly low enough to misclassify at any time. 

\begin{figure}
\centering
  \begin{minipage}[b]{0.4\linewidth}
    \includegraphics[width = \linewidth,height=3cm]{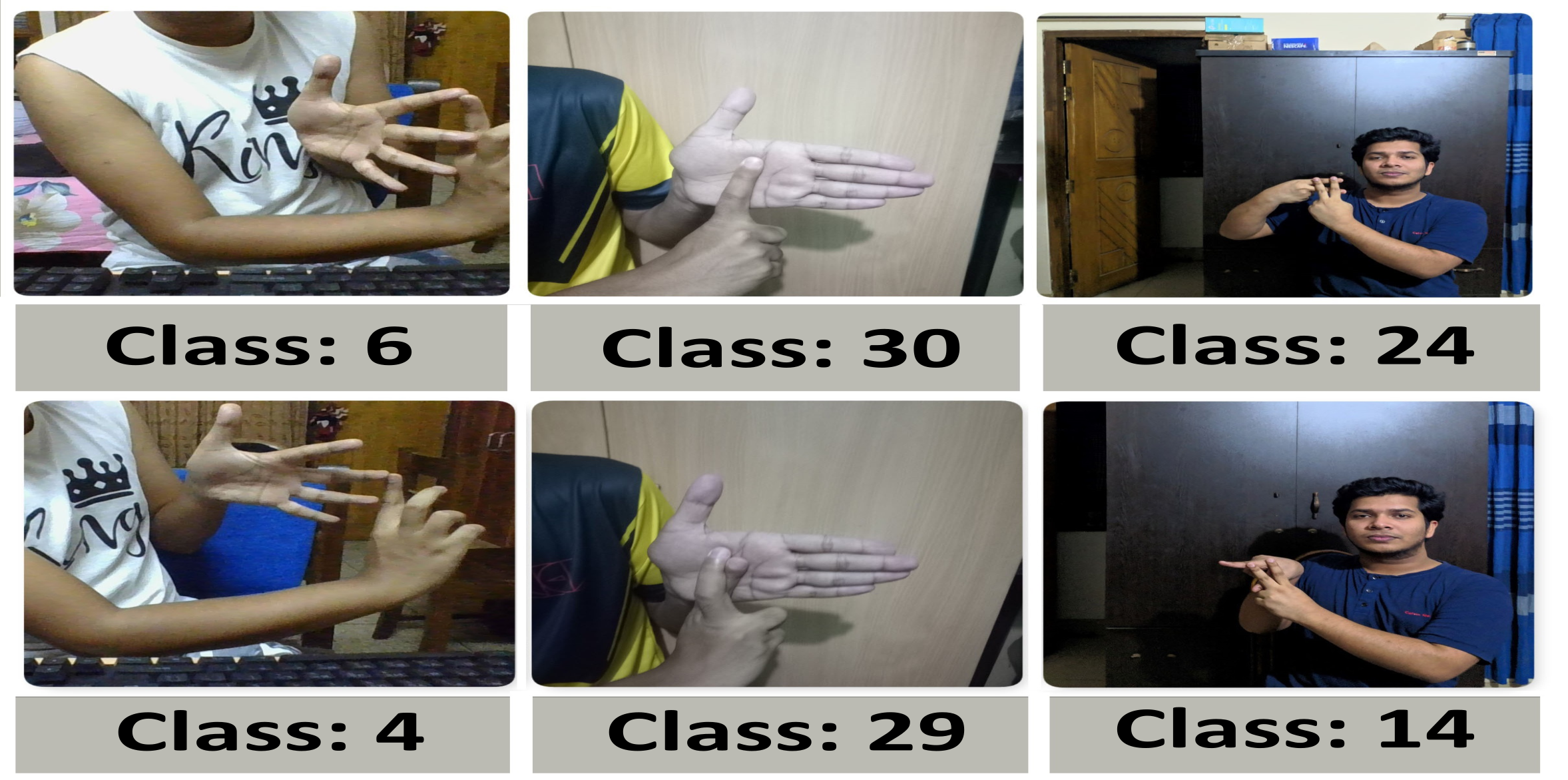}
    \caption{Signs in BdSL36 letters with appearance similarity. We can see from the figure that class 4,6, class 14,24 and class 29,30 have high similarities in appearance. These signs are mostly misclassified by the classifiers.} 
    \label{similar} 
  \end{minipage} 
  \quad
  \begin{minipage}[b]{0.4\linewidth}
    \includegraphics[width = \linewidth,height=3cm]{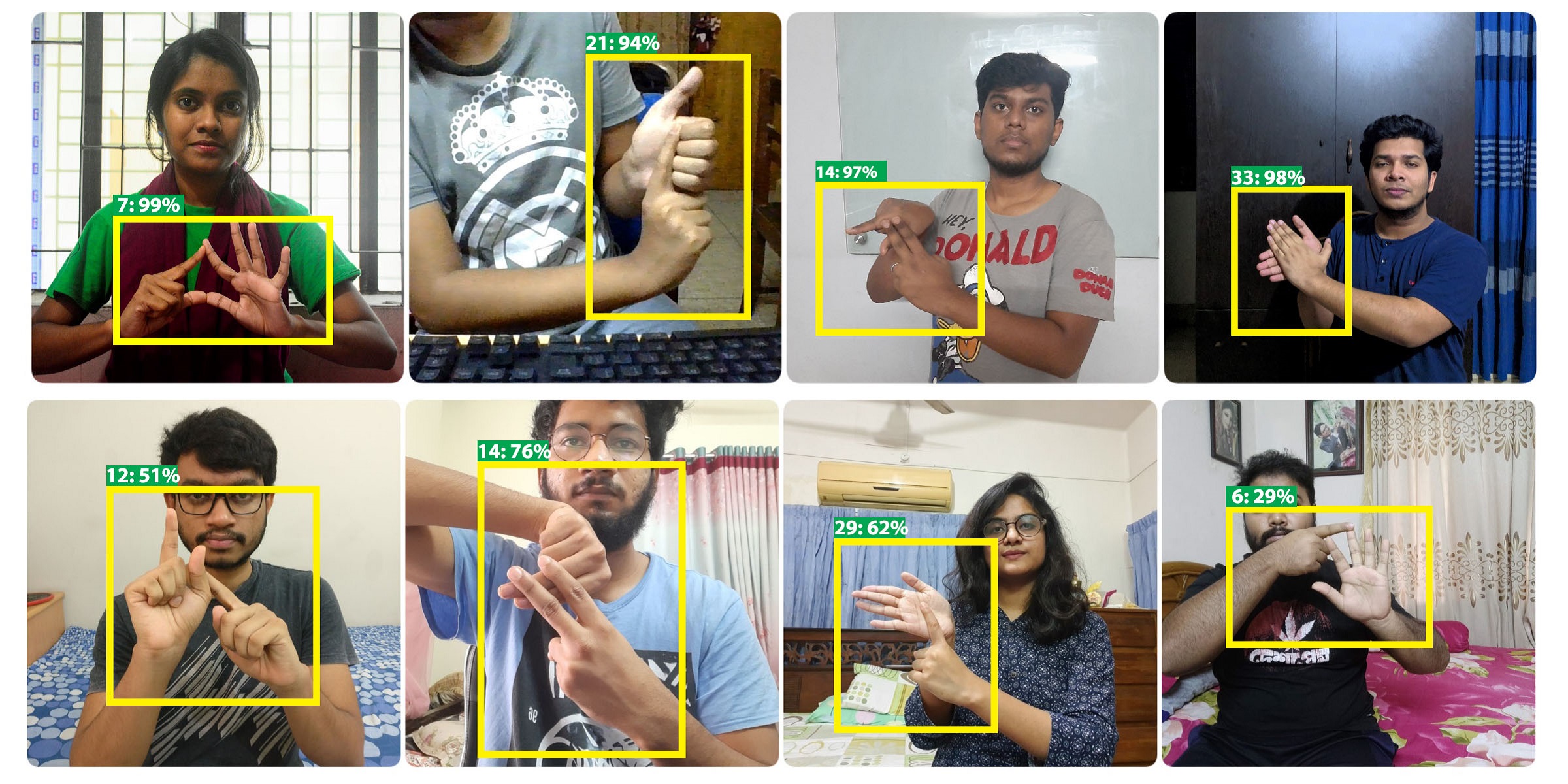}
    \caption{Sample detection results on the BdSL36v3 dataset. The top row shows the correctly classified images, where the bottom row shows the images which are correctly detected but wrongly classified.}
    \label{detection} 
  \end{minipage} 

\end{figure}

On the other hand, the user-level evaluation shows that, despite a high accuracy on the test set, classifier 2 does not perform well for six classes. Users report these classes to need multiple captures and predict with a less confidence rate. In fig. \ref{similar}, we can see that class 24, and class 14 are incredibly similar and class 24 has got the worst performance with less than 55\% confidence rate and less than 40\% accuracy with multiple tries. Users report shows that class 24 is mostly misclassified as 14, class 4 is misclassified as 3, and classes 29 and 30 are misclassified between each other. Consequently, as illustrated in fig. \ref{detection},  BdSL letters recognition also brings challenges to the detection task. Though the target is detected accurately, some of the signs get misclassified.

\section{Conclusion}
In this work, we build a large-scale dataset, named BdSL36 with different versions, for BdSL letters recognition, detection and show the usefulness of background augmentation. Our dataset includes over four million images of 36 BdSL letters. Compared with previous datasets, all version of the BdSL36 conforms to several characteristics of real environments which is suitable for building a real-time interpreter. Moreover, we also evaluate several state-of-the-art recognition and detection methods on our dataset. The results demonstrate that it is possible to generate a classifier with a small scale dataset with proper tuning of the datasets. We hope this work will help advance future research in the field of BdSL and also for other sign languages.


\bibliographystyle{splncs}
\bibliography{egbib}

\end{document}